\documentclass[a4paper,conference]{IEEEtran}
\usepackage{algpseudocode, algorithm}
\usepackage{etoolbox}\AtBeginEnvironment{algorithmic}{\small}
\usepackage{multirow}
\usepackage{graphicx,color}
\usepackage[savepos]{zref}
\usepackage{amsmath}
\usepackage{environ} 
\usepackage{mathtools,bm}
\usepackage{soul}
\usepackage{cancel}
\usepackage{booktabs}
\usepackage{caption} 
\usepackage{subfig}
\usepackage[
  separate-uncertainty = true,
  multi-part-units = repeat
]{siunitx}

\def\endthebibliography{%
  \def\@noitemerr{\@latex@warning{Empty `thebibliography' environment}}%
  \endlist
}

\begin{document}

\title{MaxDropout: Deep Neural Network Regularization Based on Maximum Output Values}

\author{\IEEEauthorblockN{Claudio Filipi Goncalves do Santos}
\IEEEauthorblockA{
Federal University of Sao Carlos -\\ UFSCar, Brazil\\
cfsantos@ufscar.br}
\and
\IEEEauthorblockN{Danilo Colombo}
\IEEEauthorblockA{Petr\'oleo Brasileiro - Petrobras\\
colombo.danilo@petrobras.com.br}
\and
\IEEEauthorblockN{Mateus Roder\\ and Jo\~ao Paulo Papa}
\IEEEauthorblockA{S\~ao Paulo State University - \\ UNESP, Brazil\\
\{mateus.roder,joao.papa\}@unesp.br}}

\maketitle
\begin{abstract}
Different techniques have emerged in the deep learning scenario, such as Convolutional Neural Networks, Deep Belief Networks, and Long Short-Term Memory Networks, to cite a few. In lockstep, regularization methods, which aim to prevent overfitting by penalizing the weight connections, or turning off some units, have been widely studied either. In this paper, we present a novel approach called MaxDropout, a regularizer for deep neural network models that works in a supervised fashion by removing (shutting off) the prominent neurons (i.e., most active) in each hidden layer. The model forces fewer activated units to learn more representative information, thus providing sparsity. Regarding the experiments, we show that it is possible to improve existing neural networks and provide better results in neural networks when Dropout is replaced by MaxDropout. The proposed method was evaluated in image classification, achieving comparable results to existing regularizers, such as Cutout and RandomErasing, also improving the accuracy of neural networks that uses Dropout by replacing the existing layer by MaxDropout.
\end{abstract}

\IEEEpeerreviewmaketitle

\section{Introduction}
\label{s.introduction}

Following the advent of deeply connected systems and the new era of information, tons of data are generated every moment by different devices, such as smartphones or notebooks. A significant portion of the data can be collected from images or videos, which are usually encoded in a high-dimensional domain. Deep Learning (DL) techniques have been broadly employed in different knowledge fields, mainly due to their ability to create authentic representations of the real world, even for multimodal information. Recently, DL has emerged as a prominent area in Machine Learning, since its techniques have achieved outstanding results and established several hallmarks in a wide range of applications, such as motion tracking~\cite{doulamis2018}, action recognition~\cite{cao2016}, and human pose estimation~\cite{Toshev2014DeepPose,chen2014}, to cite a few.

Deep learning architectures such as Convolutional Neural Networks (CNNs), Deep Autoencoders, and Long Short-Term Memory Networks are powerful tools that deal with different image variations such as rotation or noise. However, their performance is highly data-dependent, which can cause some problems during training and further generalization for unseen examples. One common problem is overfitting, where the technique memorizes the data either due to the lack of information or because of too complex neural network architectures.

Such a problem is commonly handled with regularization methods, which represent a wide area of study in the scientific community. The employment of one or more of such techniques provides useful improvements in different applications. Among them, two well-known methods can be referred: (i) so-called ``Batch Normalization" and (ii) ``Dropout". The former was introduced by Ioffe et al.~\cite{ioffe2015batch} and performs data normalization in the output of each layer. The latter was introduced by Srivastava et al.~\cite{srivastava2014dropout}, and randomly deactivates some neurons present in each layer, thus forcing the model to be sparse.

However, dropping neurons out at random may slow down convergence during learning. To cope with this issue, we introduced an improved approach for regularizing deeper neural networks, hereinafter called ``MaxDropout"~\footnote{https://github.com/cfsantos/MaxDropout-torch}, which shuts off neurons based on their maximum activation values, i.e., the method drops the most active neurons to encourage the network to learn better and more informative features. Such an approach achieved remarkable results for the image classification task, concerning two important well-established datasets.

The remainder of this paper is presented as follows: Section~\ref{s.related} introduces the correlated works, while Section~\ref{s.proposed} presents the proposed approach. Further, Section~\ref{s.experiments} describes the methodology and datasets employed in this work. Finally, Sections~\ref{s.results} and \ref{s.conclusion} provide the experimental results and conclusions, respectively.
\section{Related Works}
\label{s.related}

Regularization methods are widely used by several deep neural networks (DNNs) and with different architectures. The main idea is to help the system to prevent the overfitting problem, which causes the data memorization instead of generalization, also allowing DNNs to achieve better results. A well-known regularization method is Batch Normalization (BN), which works by normalizing the output of a giving layer at each iteration. The original work~\cite{ioffe2015batch} showed that such a process speeds up convergence for image classification tasks. Since then, several other works~\cite{zhang2017beyond,simon2016imagenet,wang2017gated}, including the current state-of-the-art on image classification~\cite{tan2019efficientnet}, also highlighted its importance.

As previously mentioned, Dropout is one of the most employed regularization methods for DNNs. Such an approach was developed between 2012 and 2014~\cite{srivastava2014dropout}, showing significant improvements in neural network's performance for various tasks, ranging from image classification, speech recognition, and sentimental analysis. The standard Dropout works by creating, during training time, a mask that direct multiples all values of a given tensor. The values of such a mask follow the Bernoulli distribution, being $0$ with a probability $p$ and $1$ with a probability $1 - p$ (according to the original work~\cite{srivastava2014dropout}, best value for $p$ in hidden layers is $0.5$).  During training, some values will be kept while others will be changed to $0$. Visually, it means that some neurons will be deactivated while others will work normally. 

After the initial development of the standard Dropout, Wang and Manning~\cite{wang2013fast} explored different strategies for sampling since at each mini-batch a subset of input features is turned off. Such a fact highlights an interesting Dropout feature since it represents an approximation by a Markov chain executed several times during training. Since the Bernoulli distribution tends to a Normal distribution when the dimensional space is high enough, such an approximation allows Dropout to its best without sampling.

In 2015, Kingma et al.~\cite{kingma2015variational} proposed the Variational Dropout, a generalization of Gaussian Dropout in which the dropout rates are learned instead of randomly attributed. They investigated a local reparameterization approach to reduce the variance of stochastic gradients in variational Bayesian inference of a posterior over the model parameters, thus retaining parallelizability. On the other hand, in 2017, Gal et al.~\cite{gal2017concrete} proposed a new Dropout variant to reinforcement learning models. Such a method aims to improve the performance and better calibration of uncertainties once it is an intrinsic property of the Dropout. In such a field, the proposed approach allows the agent to adapt its uncertainty dynamically as more data is seen.

Later on, Molchanov et al.~\cite{molchanov2017variational} explored the Variational Dropout proposed by Kingma et al.~\cite{kingma2015variational}. The authors extended it to situations when dropout rates are unbounded, leading to very sparse solutions in fully-connected and convolutional layers. Moreover, they achieved a reduction in the number of parameters up to $280$ times on LeNet architectures, and up to $68$ times on VGG-like networks with a small decrease in accuracy. Such a fact points out the importance of sparsity for parameter reduction and performance overall improvement.

Paralleling, other regularization methods have been emerged, like the ones that change the input of the neural network. For instance, Cutout~\cite{devries2017cutout} works by literally cutting off a region of the image (by setting the values of a random region to $0$). This simple approach shows relevant results on several datasets. Another similar regularizer is the RandomErasing~\cite{zhong2020random}, that works in the same manner, but instead of setting the values of the region to $0$, it changes these pixels to random values.

By bringing the concepts mentioned above and works close to the proposed approach, one can point out that the MaxDropout is similar to the standard Dropout, however, instead of randomly dropping out neurons, our approach follows a policy for shutting off the most active cells, representing a selection of neurons that may overfit the data, or discourage the fewer actives from extracting useful information.
\section{Proposed Approach}
\label{s.proposed}

The proposed approach aims at shutting out the most activated neurons, which is responsible for inducing sparsity in the model, at the step that encourage the hidden neurons to learn more informative features and extract useful information that positively impacts the network's generalization ability. 

\begin{figure}[!ht]
  \centerline{\begin{tabular}{cc}
      \includegraphics[width=0.47\columnwidth]{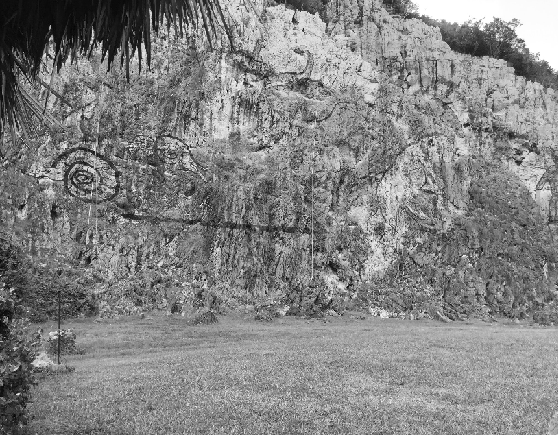} &
      \includegraphics[width=0.47\columnwidth]{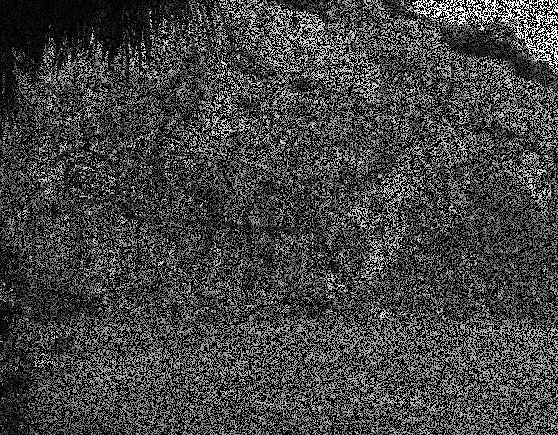} \\
      (a) & (b)\\
      \includegraphics[width=0.47\columnwidth]{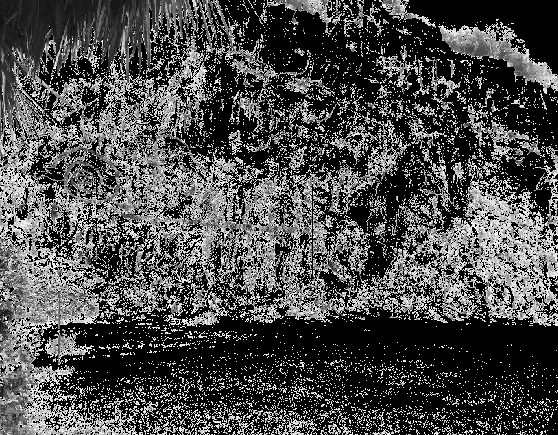} &
      \includegraphics[width=0.47\columnwidth]{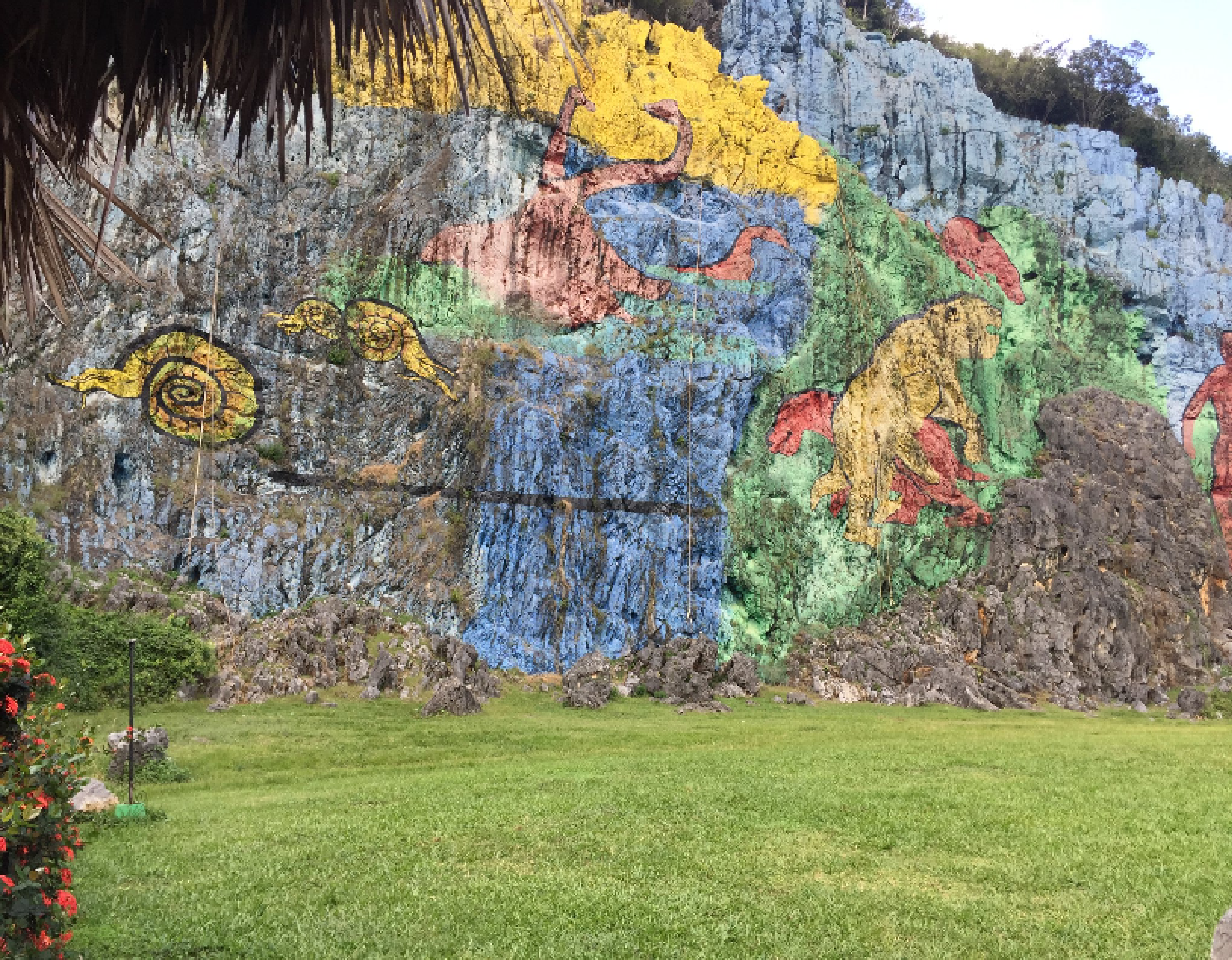} \\
      (c) & (d)\\
      \includegraphics[width=0.47\columnwidth]{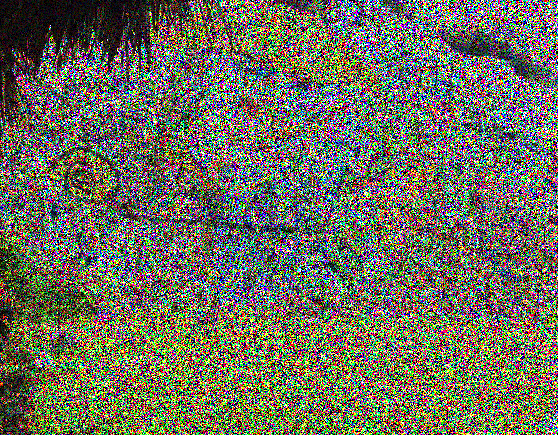} &
      \includegraphics[width=0.47\columnwidth]{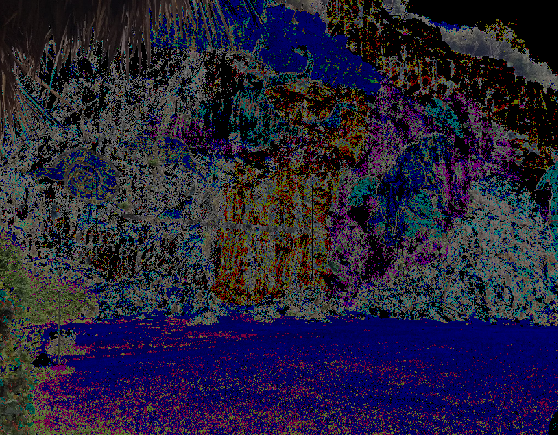} \\
      (e) & (f)\\
  \end{tabular}}
\caption{Simulation using grayscale (a)-(c) and colored images (d)-(f): (a) original grayscale image and its outcomes after (b) Dropout and (c) MaxDropout transformations, respectively, and (d) original colored image and its outcomes after (e) Dropout and (f) MaxDropout transformations, respectively. In all cases, the drop rout ate is $50$\%.}
\label{f.simulation}
\end{figure}

For the sake of visualization, Figures~\ref{f.simulation}a-c show the differences between the proposed approach and the standard Dropout, in which Figure~\ref{f.simulation}a stands for the original grayscale image and Figures~\ref{f.simulation}b and~\ref{f.simulation}c denote their corresponding outcomes after Dropout and MaxDropout. It is important to highlight that Dropout removes any pixel of the image randomly, while MaxDropout tends to inactivate the lighter pixels.

The rationale behind the proposed approach can be better visualized in a tensor-like data. Considering the colored image showed in Figure~\ref{f.simulation}d, one can observe its outcome after Dropout and MaxDropout transformations in Figures~\ref{f.simulation}e and~\ref{f.simulation}f, respectively. Regarding standard Dropout, the image looks like a colored uniform noise, while MaxDropout could remove entire regions composed of bright pixels (i.e., pixels with high activation values, as expected).

For the sake of clarification purposes, Algorithm~\ref{maxdropout-algorithm} implements the proposed MaxDropout\footnote{The pseudocode uses Keras syntax.}: the main loop in Lines $1-9$ is in charge of the training procedure, and the inner loop in Lines $2-8$ is executed for each hidden layer. Line $3$ computes a random value uniformly distributed that is going to work as the dropout rate $r$. The output of each layer produces an $x\times y\times z$ tensor, where $x$ and $y$ stand for the image's size, and $z$ denotes the number of feature maps produced for each convolutional kernel. Line $4$ creates a copy of the original tensor and uses an $L_{2}$ normalization to produce an output between $0$ and $1$.

\begin{algorithm}[h]
  \caption{Pseudocode for MaxDropout training algorithm.}
  \label{maxdropout-algorithm}
\begin{algorithmic}[1]

\While {$training$}
	\For{each layer}
		\State $rate\gets U(0, r)$
		\State $normTensor\gets L2Normalize(Tensor)$
		\State $max\gets Max(normTensor)$
		\State $keptIdx\gets IdxOf(normTensor, (1-rate)*max)$
		\State $returnTensor\gets Tensor * KeptIdx$
	\EndFor
\EndWhile
\end{algorithmic}
\end{algorithm}

Later, Line $5$ finds the biggest value in the normalized tensor, once it may not be equal to one\footnote{Depending on the floating-point precision, the maximum value can be extremely close but not equal to one.}. Line $6$ creates another tensor with the same shape as the input one and assigns $1$ where $(1 - rate) \times max$ at a certain tensor position is greater than a given threshold; otherwise it sets such a position to $0$. Finally, Line $7$ creates the tensor to be used in the training phase, where each position of the original tensor is multiplied by the value in the respective position of the tensor created in the line before. Therefore, such a procedure guarantees that only values smaller than the threshold employed in Line $3$ go further on.
\section{Experiments}
\label{s.experiments}

In this section, we describe the methodology employed to validate the robustness of the proposed approach. The hardware used for the paper is an Intel Xeon Bronze\textsuperscript{\textregistered} $3104$ CPU with $6$ cores ($12$ threads), $1.70$GHz, $96$GB RAM with $2666$Mhz, and a GPU Nvidia Tesla P$4$ with $8$GB. Since most of the regularization methods aim to improve image classification tasks, we decided to follow the same protocol and approaches for a fair comparison.

\subsection{Neural Network Structure}
\label{ss.nnstructure}

Regarding the neural network structure, we evaluated the proposed approach in two different practices. For the former experiments, regularization layers were added to a neural network that does not drop any transformation between layers. Concerning the latter experiments, the standard Dropout~\cite{srivastava2014dropout} layers were changed by the MaxDropout one to compare results.

For the first experiment, ResNet18~\cite{he2016identity} was chosen because such an architecture has been used in several works for comparison purposes when coming to new regularizer techniques. ResNet18 is compounded by a sequence of convolutional residual blocks, followed by the well-known BatchNormalization~\cite{ioffe2015batch}. As such, a MaxDropout layer was added between these blocks, changing the basic structure during training but keeping it to inference purposes.

In the second experiment, a slightly different approach has been performed. Here, a neural network that already has the Dropout regularization in its composition was considered for direct comparison among methods. The WideResNet~\cite{zagoruyko2016wide} uses Dropout layers in its blocks with outstanding results on image classification tasks, thus becoming a good choice.

\subsection{Training Protocol}
\label{ss.training}

In this work, we considered a direct comparison with other regularization algorithms. To be consistent with the literature, we provided the error rate instead of the accuracy itself~\cite{devries2017cutout, zagoruyko2016wide,zhong2020random}. Nonetheless, to ensure that the only difference between the proposed approach and the baselines used for comparison purposes concerns the MaxDropout layer, we strictly followed the protocols according to the original works.

To compare MaxDropout with other regularizers, we followed the protocol proposed by DeVries and Taylor~\cite{devries2017cutout}, in which five runs were repeated, and the mean and the standard deviation are used for comparison purposes. For the experiment, the images from the datasets were normalized per-channel using mean and standard deviation. 

During the training procedure, the images were shifted four pixels in every direction and then cropped into $32$x$32$ pixels. Besides, the images were horizontally mirrored with a $50\%$ probability. In such a case, two comparisons were provided. In the first case, besides the data augmentation already described, only the MaxDropout was included in the ResNet18 structure, directly comparing to the other methods. Regarding the second case, the Cutout data augmentation was included, providing a direct comparison of the results, showing that the proposed approach can work nicely.

As previously mentioned, to evaluate the MaxDropout against the standard Dropout, we choose the Wide Residual Network~\cite{zagoruyko2016wide}, and the same training protocol and parameters were employed to make sure the only difference concerns the type of neuron dropping.

\subsection{Datasets}
\label{ss.datasets}

\begin{figure}[!ht]
\centering
    \begin{tabular}{cc}
        \includegraphics[width=4cm]{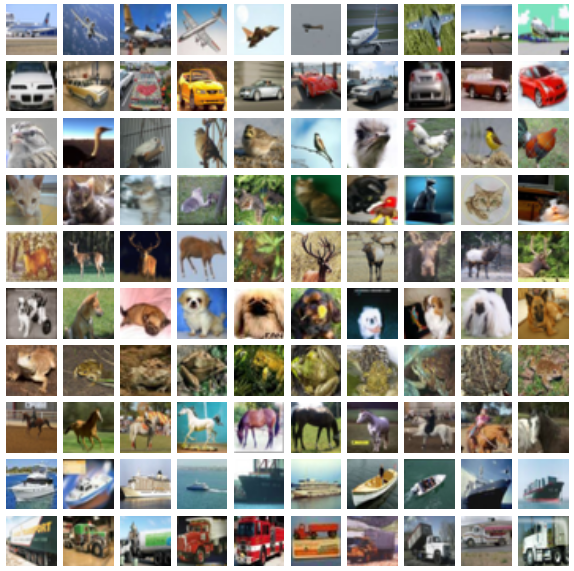}   &
        \includegraphics[width=4cm]{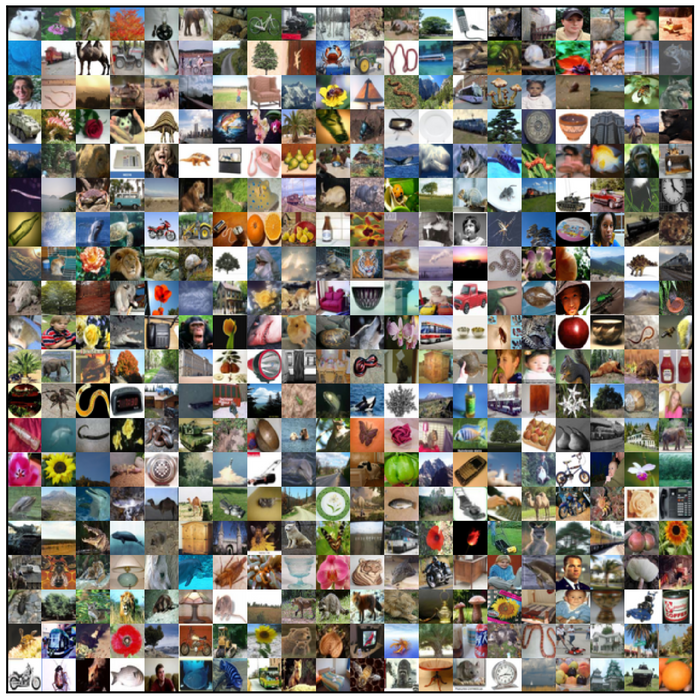} \\
        (a) & (b)
    \end{tabular}
    \caption{Random training samples from: (a) CIFAR-10 and (b) CIFAR-100 datasets.}
    \label{f.datasets}
\end{figure}

In this work, two well-established datasets in the literature were employed, i.e., CIFAR-10~\cite{Krizhevsky09learningmultiple} and its enhanced version CIFAR-100~\cite{Krizhevsky09learningmultiple}. Using such datasets allows us to compare the proposed approach toward important baseline methods, such as the standard Dropout~\cite{srivastava2014dropout} and CutOut~\cite{devries2017cutout}. Figure~\ref{f.datasets} portrays random samples extracted from the datasets mentioned above.

CIFAR-10 dataset comprises $10$ classes equally distributed in $60,000$ colored image samples, with a dimension of $32$x$32$ pixels. The entire dataset is partitioned into $50,000$ training images and $10,000$ test images. On the other hand, CIFAR-100 dataset holds similar aspects of its smaller version, but now with $100$ classes equally distributed in $60,000$ colored image samples, with 600 images samples per class. Nonetheless, the higher number of classes and the low number of samples per class make image classification significantly hard in this case.
\section{Experimental Results}
\label{s.results}

This section is divided into four main parts. First, we provided a convergence study during training for all experiments. Later, we compared the results of MaxDropout with other methods showing that, when combined with other regularizers, MaxDropout can lead to even better performance than their original versions. Finally, in the last part, we make a direct comparison between the proposed approach and standard Dropout by replacing the equivalent layer with the MaxDropout in the Wide-ResNet.

\subsection{Training Evolution}
\label{ss.training}

\begin{figure}[htb!]
	\centering                
	
	\includegraphics[width=\columnwidth]{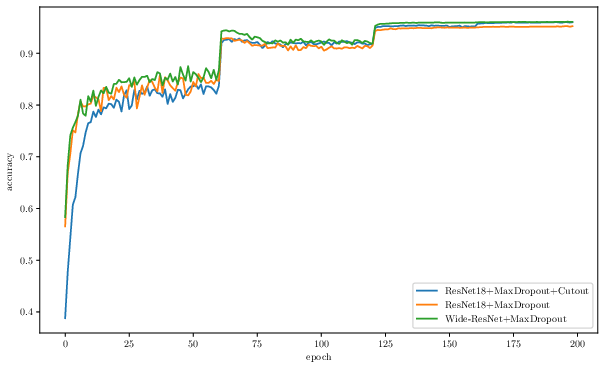}
	\caption{Convergence over CIFAR-10 test set.}
	\label{f.plot_cifar10}
\end{figure}

\begin{figure}[htb!]
	\centering                
	\includegraphics[width=\columnwidth]{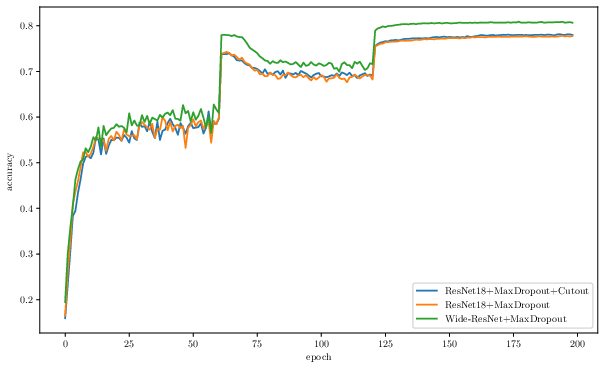}
	\caption{Convergence over CIFAR-100 test set.}
	\label{f.plot_cifar100}               %
\end{figure}

Figures~\ref{f.plot_cifar10} and~\ref{f.plot_cifar100} depict the mean accuracies concerning the test set considering the $5$ runs during training phase. Since we are dealing with regularizers, it makes sense to analyze their behavior during training and, for each epoch, compute their accuracy over the test set. One can notice that the proposed approach can improve the results even when the model is near to overfit. 

\subsection{Comparison Against Other Regularizers}
\label{ss.max}

As aforementioned, we considered a comparison against some baselines over five runs and exposed their mean accuracies and standard deviation in Table~\ref{tbl.regs}. Such results evidence the robustness of the proposed approach against two other well-known regularizers, i.e., Cutout, and the RandomErasing.

\begin{table}[htb!]
\centering
\begin{tabular}{@{} l *4c @{}}

\toprule
 \multicolumn{1}{c}{Approach}    	& CIFAR-100  & CIFAR-10  \\ 
\midrule
 ResNet18~\cite{he2016identity,zhong2020random} 		& ${24.50 \pm 0.19}$ & ${5.17 \pm 0.18}$ &  \\ 
 ResNet18+RandomErasing~\cite{zhong2020random} 			& ${24.03 \pm 0.19}$ & ${4.31 \pm 0.07}$ &  \\
 ResNet18+Cutout~\cite{devries2017cutout}  			& ${21.96 \pm 0.24}$ & \bm{${3.99 \pm 0.13}$} &  \\
 ResNet18+MaxDropout			& \bm{${21.93 \pm 0.07}$} & ${4.66 \pm 0.14}$ &  \\ \bottomrule
 \end{tabular}
 \caption{Results of MaxDropout and other regularizers}
 \label{tbl.regs}
\end{table}

From Table~\ref{tbl.regs}, one can notice that when MaxDropout is incorporated within ResNet18 blocks, it allows the model to accomplish relevant and better results.  Regarding the CIFAR-10 dataset, the model that uses MaxDropout achieved a reduction of around $0.5\%$ in the error rate when compared to ResNet18. However, concerning the CIFAR-100 dataset, the model achieved over $2\%$ less error than the same baseline, besides being statistically similar to Cutout.

\subsection{Working Along with Other Regularizers}
\label{ss.other_reg}

Since MaxDropout works inside the neural network by changing the hidden layers' values, it permits the concomitant functionality with other methods that change information from the input, such as Cutout. Table~\ref{tbl.cutmax} portrays the results of each stand-alone approach and their combination. From these results, one can notice a slight improvement in performance considering the CIFAR-100 dataset, but it ends up as a relevant gain on CIFAR-10 dataset, reaching the best results so far.

\begin{table}[htb!]
\centering
\begin{tabular}{@{} l *4c @{}}

\toprule
 \multicolumn{1}{c}{Regularizer}    	& CIFAR-100  & CIFAR-10  \\ 
\midrule

 Cutout~\cite{devries2017cutout} 			& ${21.96 \pm 0.24}$ & ${3.99 \pm 0.13}$ &  \\
 MaxDropout 			& ${21.93 \pm 0.07}$ & ${4.66 \pm 0.14}$ &  \\
 MaxDropout + Cutout		& \bm{${21.82 \pm 0.13}$} & \bm{${3.76 \pm 0.08}$} &  \\ \bottomrule
 \end{tabular}
 \caption{Results of the MaxDropout combined with Cutout.}
 \label{tbl.cutmax}
\end{table}

\subsection{MaxDropout x Dropout}
\label{ss.maxvsdrop}

One interesting point such a work stands for concerns the following question: Is the MaxDropout comparable to the standard Dropout~\cite{srivastava2014dropout}? To answer this question, we compared the proposed approach against standard Dropout by replacing it with MaxDropout on the Wide Residual Network (WRN).

From Table~\ref{tbl.wrn}, one can observe the model using MaxDropout works slightly better than standard Dropout, leading to dropping in the error rate regarding CIFAR-100 and CIFAR-10 datasets by $0.04$ and $0.05\%$, respectively. Although it may not look an impressive improvement, we showed that the proposed approach has a margin to improve the overall results, mainly when the threshold of the MaDropout is taken into account (i.e., ablation studies)\footnote{We did not show the standard deviation since the original original study did not present such an information as well.}.

\begin{table}[htb!]
\centering
\begin{tabular}{@{} l *4c @{}}

\toprule
 \multicolumn{1}{c}{Model}    	& CIFAR-100  & CIFAR-10  \\ 
\midrule

 WRN~\cite{zagoruyko2016wide} 				& ${19.25}$ & ${4.00}$ &  \\
 WRN + Dropout~\cite{zagoruyko2016wide} 			& ${18.85}$ & ${3.89}$ &  \\
 WRN + MaxDropout		& \textbf{18.81} & \textbf{3.84} &  \\ \bottomrule
 \end{tabular}
 \caption{Results of Dropout and MaxDropout over the WRN.}
 \label{tbl.wrn}
\end{table}
\section{Discussion}
\label{s.discuss}

Unfortunately, the approaches employed for comparison purposes did not release their training evolution for a direct comparison in Section~\ref{ss.training}. Nevertheless, it is possible to observe that all models performed very well for the image classification task. In Table~\ref{tbl.regs}, MaxDropout shows a result as good as Cutout for CIFAR-100 dataset, demonstrating it performs as expected when improving baseline models' results. However, it did not perform as well for CIFAR-10 dataset, but it still improves the baseline model results by almost $0.5\%$.

Results from Table~\ref{tbl.cutmax} show that the MaxDropout supports the improvement when another regularizer is used along with. Although Cutout has been used to demonstrate the proposed approach's effectiveness, one can consider other similar regularizers. The most interesting results can be found in Table~\ref{tbl.wrn}, where MaxDropout is directly compared to the standard Dropout. It shows relevant gains over the baseline model, and it performs a little better than Dropout using the same drop rate, indicating that it may be the case to find out the best drop rates for MaxDropout, which can be data or model-dependent.

\section{Conclusions and Future Works}
\label{s.conclusion}

In this paper, we introduced MaxDropout, an improved version of the original Dropout method. Experiments show that it can be incorporated into existing models, working along with other regularizers, such as Cutout, and can replace the standard Dropout with some accuracy improvement.

With relevant results, we intend to conduct a more in-depth investigation to figure out the best drop rates depending on the model and the training data. Moreover, the next step is to re-implement MaxDropout and make it available in other frameworks, like TensorFlow and MXNet, and test in other tasks, such as object detection and image segmentation.

Nonetheless, we showed that MaxDropout works very well for image classification tasks. For future works, we intended to perform evaluations in other different tasks such as natural language processing and automatic speech recognition.

\section*{Acknowledgments}
The authors are grateful to S\~{a}o Paulo Research Foundation - FAPESP (\#2013/07375-0, \#2014/12236-1, \#2017/25908-6, and \#2019/07825-1, Brazilian National Council for Scientific and Technological Development - CNPq (\#307066/2017-7, and \#427968/2018-6), and Petrobras (\#2017/00285-6).

\bibliographystyle{IEEEtran}
\bibliography{references}

\end{document}